\documentclass[runningheads]{llncs}
\usepackage{float}
\usepackage{booktabs}
\usepackage{array}
\usepackage{multirow}
\newcolumntype{C}[1]{>{\centering\arraybackslash}p{#1}}
\usepackage{longtable}
\usepackage{amsmath}
\usepackage{hyperref}
\usepackage{overpic}
\usepackage{subcaption}
\usepackage{comment}
\usepackage{graphicx}
\usepackage{marvosym}
\begin{document}
\title{UrFound: Towards Universal Retinal Foundation Models via Knowledge-Guided Masked Modeling}
\titlerunning{UrFound: Universal Retinal Foundation Models}
\author{Kai Yu\inst{1} \and
Yang Zhou\inst{1}\textsuperscript{(\Letter)} \and 
Yang Bai\inst{1} \and 
Zhi Da Soh\inst{2} \and Xinxing Xu\inst{1} \and Rick Siow Mong Goh\inst{1} \and Ching-Yu Cheng\inst{2} \and Yong Liu\inst{1}
}
\authorrunning{K. Yu et al.}
%
\institute{Institute of High Performance Computing (IHPC), Agency for Science, Technology and Research (A*STAR), 1 Fusionopolis Way, \#16-16 Connexis, 138632, Singapore, Republic of Singapore\\ \email{zhou\_yang@ihpc.a-star.edu.sg}\\ \and Singapore Eye Research Institute, Singapore National Eye Centre\\ \and Centre for Innovation and Precision Eye Health, National University of Singapore, Singapore}
\maketitle              

\begingroup
    \renewcommand{\thefootnote}{}
    \footnotetext{Yang Zhou—Corresponding Author.}
    \addtocounter{footnote}{-1}
\endgroup

\begin{abstract}
Retinal foundation models aim to learn generalizable representations from diverse retinal images, facilitating label-efficient model adaptation across various ophthalmic tasks. 
Despite their success, current retinal foundation models are generally restricted to a single imaging modality, such as Color Fundus Photography (CFP) or Optical Coherence Tomography (OCT), limiting their versatility. 
Moreover, these models may struggle to fully leverage expert annotations and overlook the valuable domain knowledge essential for domain-specific representation learning.
To overcome these limitations, we introduce UrFound, a retinal foundation model designed to learn universal representations from both multimodal retinal images and domain knowledge. 
UrFound is equipped with a modality-agnostic image encoder and accepts either CFP or OCT images as inputs. 
To integrate domain knowledge into representation learning, we encode expert annotation in text supervision and propose a knowledge-guided masked modeling strategy for model pre-training. It involves reconstructing randomly masked patches of retinal images while predicting masked text tokens conditioned on the corresponding retinal image. 
This approach aligns multimodal images and textual expert annotations within a unified latent space, facilitating generalizable and domain-specific representation learning.
Experimental results demonstrate that UrFound exhibits strong generalization ability and data efficiency when adapting to various tasks in retinal image analysis.
By training on $\sim$180k retinal images, UrFound significantly outperforms the state-of-the-art retinal foundation model trained on up to 1.6 million unlabelled images across 8 public retinal datasets. 
Our code and data are available at \url{https://github.com/yukkai/UrFound}.

\end{abstract}
\section{Introduction}
Foundation models (FMs) are large, powerful artificial intelligence (AI) models pre-trained on vast amounts of unlabeled data. By learning fundamental patterns and relationships within diverse data, FMs gain the ability to adapt to diverse downstream tasks with minimal additional training~\cite{zhou2023foundation}. Notable examples of FMs, such as CLIP~\cite{radford2021learning}, SAM~\cite{Kirillov_2023_ICCV}, and GPT4~\cite{OpenAI2023ChatGPT}, have demonstrated impressive generalization capabilities in various real-world scenarios, including image classification, image segmentation, and natural language processing.

Medical FMs are a specialized type of FM designed for the medical domain~\cite{ma2024segment,moor2023foundation,soenksen2022integrated,tiu2022expert,zhou2023advancing}, representing one of the most notable advancements in medical AI. Among these, Medical Vision-Language pre-training stands out as a specific solution that improves medical image analysis by learning domain-specific features from medical images paired with corresponding clinical descriptions or reports~\cite{zhang2022contrastive,li2024llava,chen2023towards,bazi2023vision}. 
Recent medical FMs have focused heavily on radiology, particularly chest X-rays~\cite{you2023cxr,zhang2023knowledge,liu2023utilizing}. For retinal FMs, RETFound~\cite{zhou2023foundation} has been proposed, which is pre-trained on 1.6 million retinal images using Masked Autoencoders (MAE). Another notable example is FLAIR~\cite{silva2023foundation}, a vision-language model that leverages the CLIP architecture to enhance performance in retinal imaging analysis, supporting zero-shot and few-shot inference through text supervision.
Unlike task-specific models that may yield suboptimal results in the presence of domain shifts, retinal FMs demonstrate robust generalization capabilities across different retinal datasets and tasks. This presents an attractive solution to enhance model efficacy and reduce the annotation burden on experts, thereby enabling widespread clinical AI applications in retinal imaging.

Albeit impressive, existing retinal FMs are restricted to processing a single imaging modality, such as Colour Fundus Photography (CFP) and Optical Coherence Tomography (OCT). In clinical ophthalmology, diagnosis often involves multiple modalities, including CFP, OCT, and Fundus Fluorescence Angiography (FFA) images. This requires training separate FMs for each modality, resulting in higher maintenance costs and hindering the acquisition of complementary information across modalities. The question arises: Can a retinal FM be developed to process multiple modalities? Moreover, expert domain knowledge, often in the form of labels or medical reports, is crucial for effective retinal image analysis. It guides models in capturing clinically relevant information, ensuring clinical significance in real-world healthcare scenarios. However, current retinal FMs struggle to fully leverage expert annotations, potentially hindering specialized representation learning. Another question arises: Can domain knowledge be incorporated into a retinal FM for better generalization ability?

To address the research problems mentioned above, we introduce UrFound, a universal retinal FM designed to learn versatile representations from both multimodal retinal images and domain knowledge. UrFound employs a modality-agnostic image encoder for processing CFP or OCT images and integrates domain knowledge from categorical labels and clinical descriptions through text supervision. To achieve this, we convert expert annotations into detailed clinical descriptions and propose a knowledge-guided masked modeling strategy for UrFound pre-training. This strategy includes a masked image modeling branch to reconstruct randomly masked patches of retinal images, and a conditional masked language modeling branch to predict masked word tokens based on the corresponding retinal image. This approach aligns multimodal images and textual expert annotations within a unified latent space, facilitating domain-specific representation learning.

Empirically, we find that incorporating domain knowledge into the retinal FM through text supervision enhances generalization ability. Furthermore, UrFound captures information from both CFP and OCT images and performs well with both imaging modalities. Despite being pre-trained on a relatively small dataset of 180k retinal images with expert annotations, UrFound significantly outperforms state-of-the-art retinal FMs trained on up to 1.6 million unlabeled images across eight public retinal datasets. This demonstrates the effectiveness of multimodal images and domain knowledge in training powerful retinal FMs.

Our contribution is threefold: 1. We propose UrFound, a universal retinal foundation model capable of processing CFP and OCT images while incorporating domain knowledge from expert annotations. 2. We introduce the knowledge-guided masked modeling strategy that unifies the pre-training from multimodal images and clinical descriptions, integrating domain knowledge effectively. 3. We provide comprehensive evaluations, comparing UrFound against state-of-the-art retinal FMs across eight public retinal datasets.

\section{The UrFound Model}
\begin{figure}[t]
\centering
\includegraphics[width=0.9\textwidth]{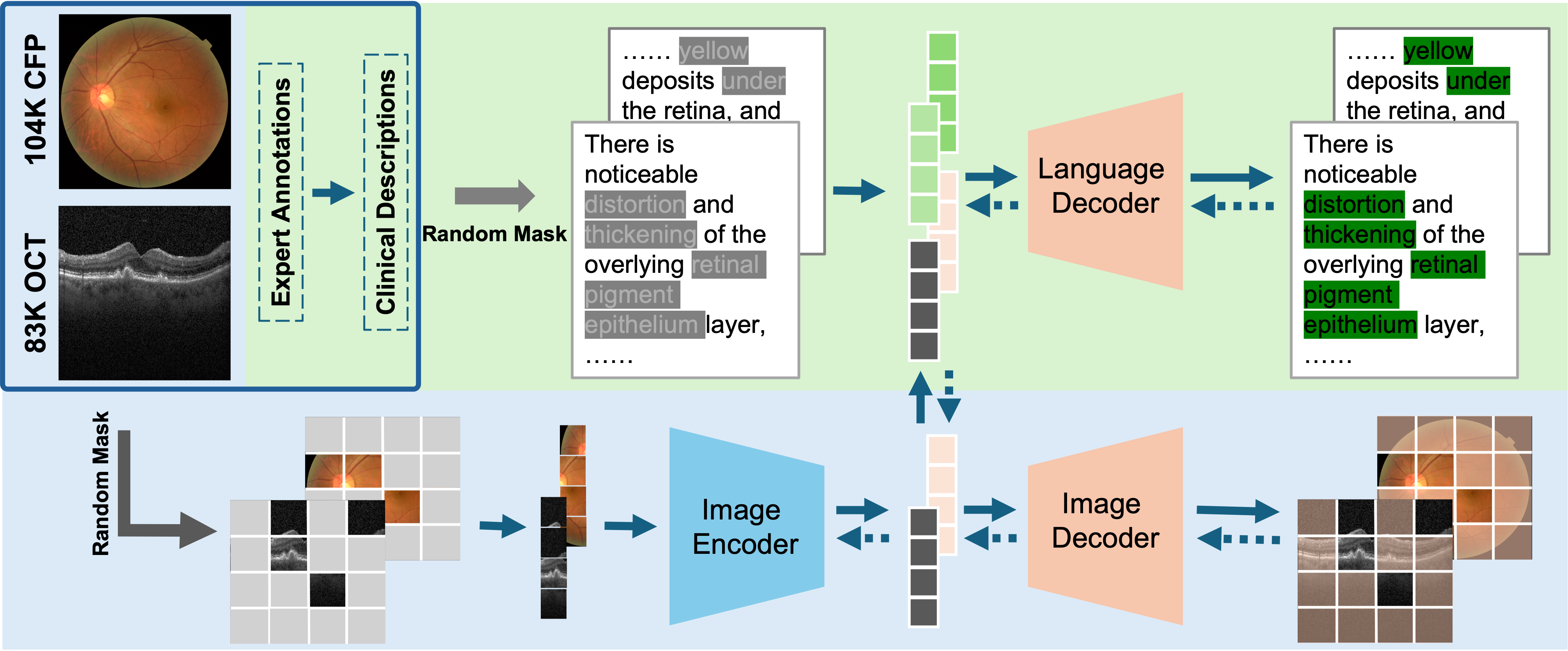} 
    \put(-77,15){\tiny{$\mathcal{L}_{MIM}$}}
    \put(-77,81){\tiny{$\mathcal{L}_{MLM}$}}

\caption{Knowledge-guided masked modeling framework for UrFound pre-training. Solid arrows represent data flow, while dashed arrows indicate gradient flow.}
\label{fig:framework}
\vskip -15pt
\end{figure}

In this section, we propose UrFound, a retinal FM designed for CFP and OCT images, as the initial step toward developing universal retinal FMs. UrFound is trained with guidance from expert annotations, which can take the form of categorical labels, clinical descriptions, or any other formats that can be encoded in text supervision. UrFound aims to learn domain-specific representations by reconstructing masked patches of a retinal image while predicting masked word tokens of textual domain knowledge conditioned on the unmasked image patches. 

Fig.~\ref{fig:framework} provides an overview of the UrFound model. UrFound has an image encoder that learns the latent representation of retinal images as well as two decoders that reconstruct the original retinal image and predict the word tokens of the associated clinical descriptions from the latent representation, respectively. The input of the image encoder can be either a CFP or OCT image. During pre-training, we apply masked image modeling to randomly mask certain patches of the input image. Then, the rest unmasked image patches are fed into the image encoder to obtain their embeddings. These embeddings are then forwarded to the image decoder to reconstruct the masked image patches, aiding the model in capturing versatile and informative visual features.

Similar to masked image modeling, we apply conditional masked language modeling to replace certain portions of word tokens of the clinical descriptions with the mask token. The language decoder is then trained to predict the original identity of the masked tokens based on both the unmasked words and the latent image representation from the image encoder. This approach encourages the model to recognize and comprehend the relationships between the retinal image and fine-grained medical knowledge. It serves to bridge the gap between visual features and textual information, integrating domain knowledge from the descriptions into the latent image representation.

\subsection{Knowledge-guided Masked Modeling} 
Formally, given a retinal image $X$, it is first reshaped into $n$ patches with the patch size $s$ (e.g., $16 \times 16$ in ViT~\cite{alexey-etal-2021-ViT}). A random mask $M \in \{0,1\}^{n}$ is generated with the mask ratio $\rho$ where 1 indicates a masked patch and 0 indicates an unmasked patch. The masked image $\tilde{X}$ is obtained as: $\tilde{X}_i = X_i \cdot (1 - M_i) + X_0 \cdot M_i, \forall i \in \{1, \cdots, n\}$, where $X_0$ represents the image [MASK] token. Let $f(\cdot)$ be the image encoder that maps each image patch to a latent representation $\mathbf{z}_i = f(\tilde{X}_i)$, and $g^{v}(\cdot)$ be the image decoder that reconstructs the original image patch $X_i$ from the latent representation. Then the mask imaging modeling (MIM) can be achieved by minimizing the following mean square error (MSE) loss:
\begin{equation}
    \mathcal{L}_{MIM} = \sum_{i=1}^{n} M_i \cdot || X_i - g^{v}(\mathbf{z}_i) ||^2_2,
\end{equation}
which measures the differences between the reconstructed and original image patches. We adopt the high-resolution trick in~\cite{zhou2023advancing} to let the model reconstruct high-resolution patches at 2× the input resolution, which allows the model to learn more detailed local features. 

For conditional masked language modeling (MLM), the input text is transformed into a sequence of tokens $W = [ w_1, \cdots, w_L ]$, where $L$ is the sequence length. Then, a certain percentage of tokens in the sequence are randomly replaced with a special [MASK] token, leading to a masked set $\mathcal{W}_{\mathcal{M}}$ and an unmasked set $\mathcal{W}_{\mathcal{N}}$. Let $\mathbf{z}$ be the average pooling of the unmasked image patch representations, and $h(\cdot)$ be the text decoder to restore the masked text tokens. The objective of MLM is to minimize the negative log-likelihood function as follows:
\begin{equation}
    \mathcal{L}_{MLM} = -\sum_{w_i \in \mathcal{W}_{\mathcal{M}}} \log P(h(w_i) | \{h(w_j), w_j \in \mathcal{W}_{\mathcal{N}} \}, \mathbf{z}),
\end{equation}
which predicts the original identities of those masked tokens based on the surrounding context and the latent image representation. The total pre-training objective function of the UrFound model is $\mathcal{L} = \mathcal{L}_{MIM} + \mathcal{L}_{MLM}$. After pre-training, the decoders are discarded and the encoder $f(\cdot)$ can be fine-tuned with a small number of data for specific downstream tasks for retinal image analysis.

\subsection{Text Preparation}
For retinal images, the majority of publicly available expert annotations come in the form of categorical labels rather than text. To maximize the utilization of domain knowledge for pre-training, we follow FLAIR \cite{silva2023foundation} to enhance categorical image labels by augmenting relevant medical findings sourced from established knowledge bases and clinical literature. For instance, the label ``drusens" might be described as ``yellow deposits under the retina" or ``numerous uniform round yellow-white lesions". Each label may have a varying number of descriptions. During pre-training, we randomly select one of these descriptions for samples in each batch, enhancing the diversity and robustness of the text supervision.

\subsection{Multimodal Image Processing}
UrFound directly learns representations for CFP and OCT images using a modality-agnostic encoder. We have observed that this straightforward implementation performs well and achieves superior generalization, particularly when training is guided by domain knowledge through masked modeling. We also explore variants that use separate patch embedding layers, encoders, and decoders for CFP and OCT imaging, respectively, while such modifications do not lead to better results in our experiments. 

\section{Experiments}
In this section, we assess the performance of UrFound compared to the state-of-the-art retinal FMs, and conduct comprehensive experiments to address the following key questions: \textbf{Q1.} Can the imaging modalities of CFP and OCT be encoded in a universal FM? \textbf{Q2.} Does domain knowledge improve the generalization ability of FMs? \textbf{Q3.} Do CFP and OCT images contain supplementary information that helps representation learning? \textbf{Q4.} How well do retinal FMs perform in terms of data efficiency? \textbf{Q5.} How effective are retinal FMs in adapting to downstream tasks compared with task-specific models?

\subsection{Experimental Setup}
We assess the capabilities of UrFound in adapting to diagnostic classification tasks with minimal additional training. In line with common practice, we add a linear classifier head on top of the learned image encoder and then fine-tune both the encoder and classifier with task-specific labels. We compare the proposed UrFound against the MAE model pre-trained on natural images as well as state-of-the-art retinal FMs including RETFound~\cite{zhou2023foundation} and FLAIR~\cite{silva2023foundation}. For these compared models, we use official checkpoints for fine-tuning. We report the area under the receiver operating curve (ROC) and the area under the precision-recall curve (PRC) as evaluation metrics. And we choose the best checkpoints with the highest ROC scores on the validation set for final evaluation.

\textbf{Datasets.}
For pre-training, we construct a training set by collecting 25 CFP datasets and one large OCT dataset, which include 103,786 CFP images and 83,484 OCT images with expert annotations, covering a wide range of ophthalmic diseases. 
We follow \cite{silva2023foundation} to augment domain knowledge and transform categorical labels into textual inputs. For the evaluation of fine-tuning performance, we test 8 publicly available datasets across three diagnostic classification tasks including diabetic retinopathy grading (IDRID~\cite{porwal2018indian}, MESSIDOR~\cite{decenciere2014feedback}, APTOS~\cite{aptos2019-blindness-detection}), glaucoma detection (PAPILA~\cite{kovalyk2022papila}, GF~\cite{ahn2018deep}), and multi-disease diagnosis (JSIEC~\cite{cen2021automatic}, Retina, OCTID~\cite{gholami2020octid}). 

\begin{table}[t]
\centering
\caption{Performance of retinal FMs on different dataset (\textbf{best}, \underline{second best}).}
\label{tab:sota_compare}
{\scriptsize
\begin{tabular}{l|C{0.8cm}C{0.8cm}|C{0.8cm}C{0.8cm}|C{0.8cm}C{0.8cm}|C{1.0cm}C{1.0cm}|lC{1.0cm}C{1.0cm}}
\toprule
\multirow{2}{*}{\textbf{Dataset}} & \multicolumn{2}{c|}{\textbf{MAE}} & \multicolumn{2}{c|}{\textbf{FLAIR}} & \multicolumn{2}{c|}{\textbf{RETFd-CFP}} & \multicolumn{2}{c|}{\textbf{RETFd-OCT}} & \multicolumn{2}{c}{\textbf{UrFound}} \\

& {ROC} & {PRC} & {ROC} & {PRC} & {ROC} & {PRC} & {ROC} & {PRC} & {ROC} & {PRC} \\

\midrule
\textbf{APTOS}       & 94.06 & 67.59 & 92.68 & 62.20 & \underline{94.26} & \textbf{71.87} & 87.56 & 53.76 & \textbf{94.86} & \underline{71.64} \\
\textbf{IDRID}       & 79.24 & 43.65 & 80.88 & 49.32 & \underline{83.33} & \underline{51.13} & 59.29 & 28.66 & \textbf{85.22} & \textbf{57.73} \\
\textbf{Messidor}    & 84.21 & 48.76 & 81.88 & 48.32 & \underline{86.40} & \underline{58.59} & 65.89 & 28.59 & \textbf{88.22} & \textbf{60.78} \\
\midrule
\textbf{PAPILA}      & 62.85 & 47.48 & \underline{74.80} & \underline{59.30} & 74.36 & 57.27 & 51.67 & 35.03 & \textbf{78.32} & \textbf{62.54} \\
\textbf{GF}          & 93.09 & 83.17 & 78.87 & 59.60 & \underline{95.68} & \textbf{88.18} & 87.61 & 70.85 & \textbf{95.75} & \underline{88.01} \\
\midrule
\textbf{JSIEC}       & 98.46 & 81.78 & 93.53 & 52.65 & \underline{99.39} & \underline{86.95} & 88.44 & 41.09 & \textbf{99.51} & \textbf{92.34} \\
\textbf{Retina}       & 74.22 & 53.70 & 77.75 & 55.33 & \underline{86.22} & \underline{71.59} & 75.43 & 53.76 & \textbf{90.09} & \textbf{79.30} \\
\midrule
\textbf{OCTID}       & 98.67 & 95.35 & 84.52 & 60.20 & 93.85 & 82.09 & \underline{99.39} & \underline{97.58} & \textbf{99.55} & \textbf{97.97} \\
\bottomrule
\end{tabular}
}
\end{table}

\textbf{Implementation details.}
We implement UrFound by using PyTorch on a single NVIDIA A100 GPU. We employ a Vision Transformer (ViT-base) with 12 Transformer blocks and a patch embedding layer as the retinal image encoder. We utilize 8 and 6 Transformer blocks as the image and text decoders, respectively. In the pre-training stage, we initialize UrFound with the MAE model and use the tokenizer of BERT-Base~\cite{devlin-etal-2019-bert} to convert clinical descriptions into word tokens. We use a mask ratio of 0.75 for image modeling and 0.5 for language modeling.
We resize the input image to 224×224 both in the pre-training stage and fine-tuning stage. Random horizontal flip and random crop are implemented in the pre-training stage. And random horizontal flip, and color jitter for data augmentation in the fine-tuning stage, each with a probability of 0.5.
The total training epoch is set to 200 with a warm-up period of 40 epochs. The learning rate is set to 1.5e-4, and the batch size is set to 128. In the fine-tuning stage, the learning rate is adjusted to 1e-4, the batch size is reduced to 16, and the training epoch is set to 50 with a warm-up period of 10 epochs.

\subsection{Main Results}
Table \ref{tab:sota_compare} shows the classification results of the compared retinal FMs fine-tuned for various retinal disease diagnosis tasks. And our results show that UrFound performs similarly to the second-best method on IDRID and JSIEC, and significantly better on the other six datasets. It can be observed that retinal FMs such as RETFound-CFP and UrFound achieve significantly better results than MAE in all the cases, which demonstrates the effectiveness of retinal FMs in learning generalizable representations for retinal imaging analysis. UrFound consistently outperforms the second best method, RETFound. This superiority can be attributed to the integrated domain knowledge in UrFound through text supervision. 
In contrast, although FLAIR also leverages domain knowledge, it does not perform well and lags behind MAE in some cases. This is possibly because FLAIR focuses on image-text alignment rather than capturing the visual features of retinal images. It results in a sub-optimal image encoder for image understanding in the pretrain-finetune setting. 

RETFound-CFP and FLAIR are designed specifically for CFP images, exhibiting subpar performance when applied to OCT images. Similarly, RETFound-OCT yields the poorest results on CFP datasets. In contrast, UrFound showcases its superiority in processing both CFP and OCT modalities. It achieves this by learning universal and comprehensive representations that span across modalities, demonstrating its capability to effectively handle diverse imaging types.

\begin{table}[t]
\centering
\caption{Performance of UrFound and its variants (\textbf{best}, \underline{second best}).}
\label{tab:ablation_mrm}
{\scriptsize
\begin{tabular}{l|C{0.8cm}C{0.8cm}|C{0.8cm}C{0.8cm}|C{0.8cm}C{0.8cm}|C{0.8cm}C{0.8cm}|C{0.8cm}C{0.8cm}|C{0.8cm}C{0.8cm}}
\toprule
\multirow{3}{*}{\textbf{Dataset}} & \multicolumn{6}{c|}{\textbf{W/O Text Supervision}} & \multicolumn{6}{c}{\textbf{W/ Text Supervision}} \\
& \multicolumn{2}{c|}{\textbf{CFP}} & \multicolumn{2}{c|}{\textbf{OCT}} & \multicolumn{2}{c|}{\textbf{CFP+OCT}} & \multicolumn{2}{c|}{\textbf{CFP}} & \multicolumn{2}{c|}{\textbf{OCT}} & \multicolumn{2}{c}{\textbf{CFP+OCT}} \\

& {ROC} & {PRC} & {ROC} & {PRC} & {ROC} & {PRC} & {ROC} & {PRC} & {ROC} & {PRC} & {ROC} & {PRC} \\

\midrule
\textbf{APTOS}       & 93.81 & 65.93 & 89.92 & 56.11 & 94.01 & 67.58 & \underline{94.36} & \underline{68.68} & 90.40 & 56.38 & \textbf{94.86} & \textbf{71.64} \\
\textbf{IDRID}       & 79.47 & 45.44 & 69.10 & 35.32 & 79.51 & 44.59 & \underline{84.64} & \underline{55.46} & 66.28 & 31.09 & \textbf{85.22} & \textbf{57.73} \\
\textbf{Messidor}    & 84.76 & 52.86 & 69.10 & 30.73 & 84.28 & 50.47 & \underline{86.24} & \underline{58.28} & 71.21 & 32.34 & \textbf{88.22} & \textbf{60.78} \\
\midrule
\textbf{PAPILA}      & 69.13 & 52.36 & 47.21 & 35.47 & 69.65 & 53.68 & \underline{73.45} & \underline{54.92} & 56.59 & 38.19 & \textbf{78.32} & \textbf{62.54} \\
\textbf{GF}          & 93.84 & 84.50 & 89.01 & 73.30 & 93.48 & 83.70 & \underline{95.16} & \underline{87.17} & 89.61 & 73.62 & \textbf{95.75} & \textbf{88.01} \\
\midrule
\textbf{JSIEC}       & 98.72 & 88.72 & 91.71 & 50.60 & 99.08 & 85.44 & \underline{99.48} & \textbf{92.84} & 92.35 & 51.33 & \textbf{99.51} & \underline{92.34} \\
\textbf{Rtina}      & 88.17 & \underline{76.01} & 70.29 & 48.78 & 87.08 & 74.34 & \underline{88.50} & 75.77 & 81.40 & 61.81 & \textbf{90.09} & \textbf{79.30} \\
\midrule
\textbf{OCTID}       & 98.40 & 95.60 & 99.37 & 97.35 & \textbf{99.59} & \underline{97.88} & 98.05 & 94.56 & 99.28 & 95.33 & \underline{99.55} & \textbf{97.97} \\
\bottomrule
\end{tabular}
}
\end{table}

\textbf{Impact of multimodal imaging and domain knowledge.}
To investigate how multimodal data and domain knowledge affect the performance of UrFound, we compared UrFound against its single-modality variants, either with or without domain knowledge. 
As shown in Table \ref{tab:ablation_mrm}, without text supervision, UrFound trained from CFP+OCT images achieves reasonably good results on both CFP and OCT datasets. This indicates that it is promising to learn universal FMs for multiple retinal imaging modalities (\textbf{Q1}). Furthermore, the inclusion of text supervision significantly enhances the performance of UrFound, which demonstrates the effectiveness of domain knowledge in learning domain-specific and generalizable representations (\textbf{Q2}). With text supervision, UrFound trained from CFP+OCT images outperforms its single-modality counterparts, which suggests that CFP and OCT images contain supplementary information beneficial for improved representation learning (\textbf{Q3}).

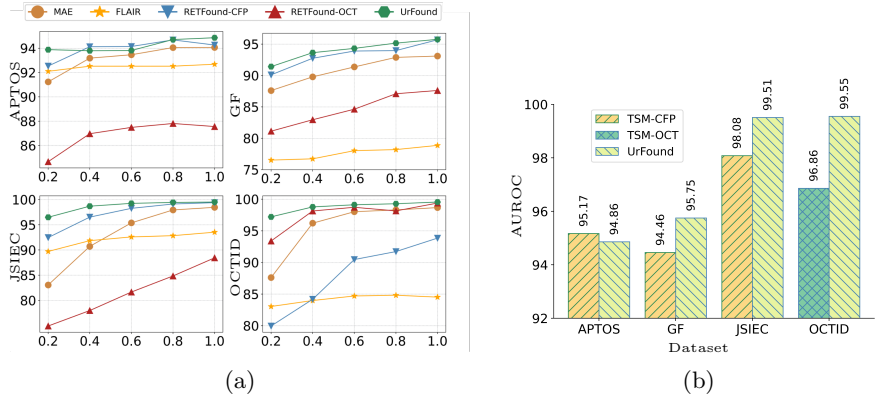
\begin{figure}[t]
    \centering
    
    \begin{subfigure}{.5\textwidth}
        \centering
        \begin{overpic}[width=1.0\textwidth]{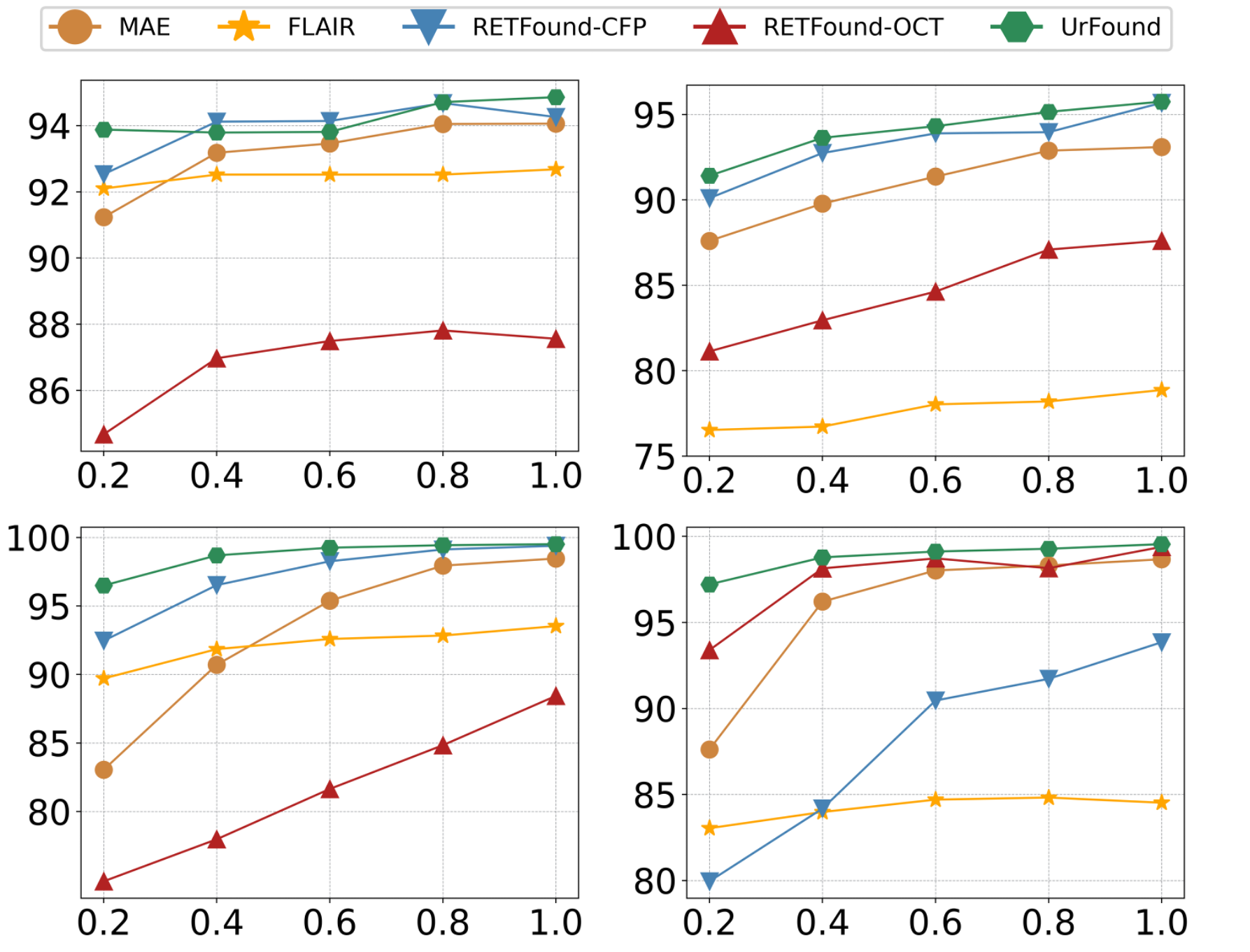}
            \put(0,50){\rotatebox{90}{\tiny{APTOS}}}
            \put(48,51){\rotatebox{90}{\tiny{GF}}}
            \put(0,15){\rotatebox{90}{\tiny{JSIEC}}}
            \put(48,12){\rotatebox{90}{\tiny{OCTID}}}
        \end{overpic}
        \caption{}
        \label{fig:sub_effi}
        
    \end{subfigure}%
    \begin{subfigure}{.5\textwidth}
        \centering
        \includegraphics[width=.8\linewidth]{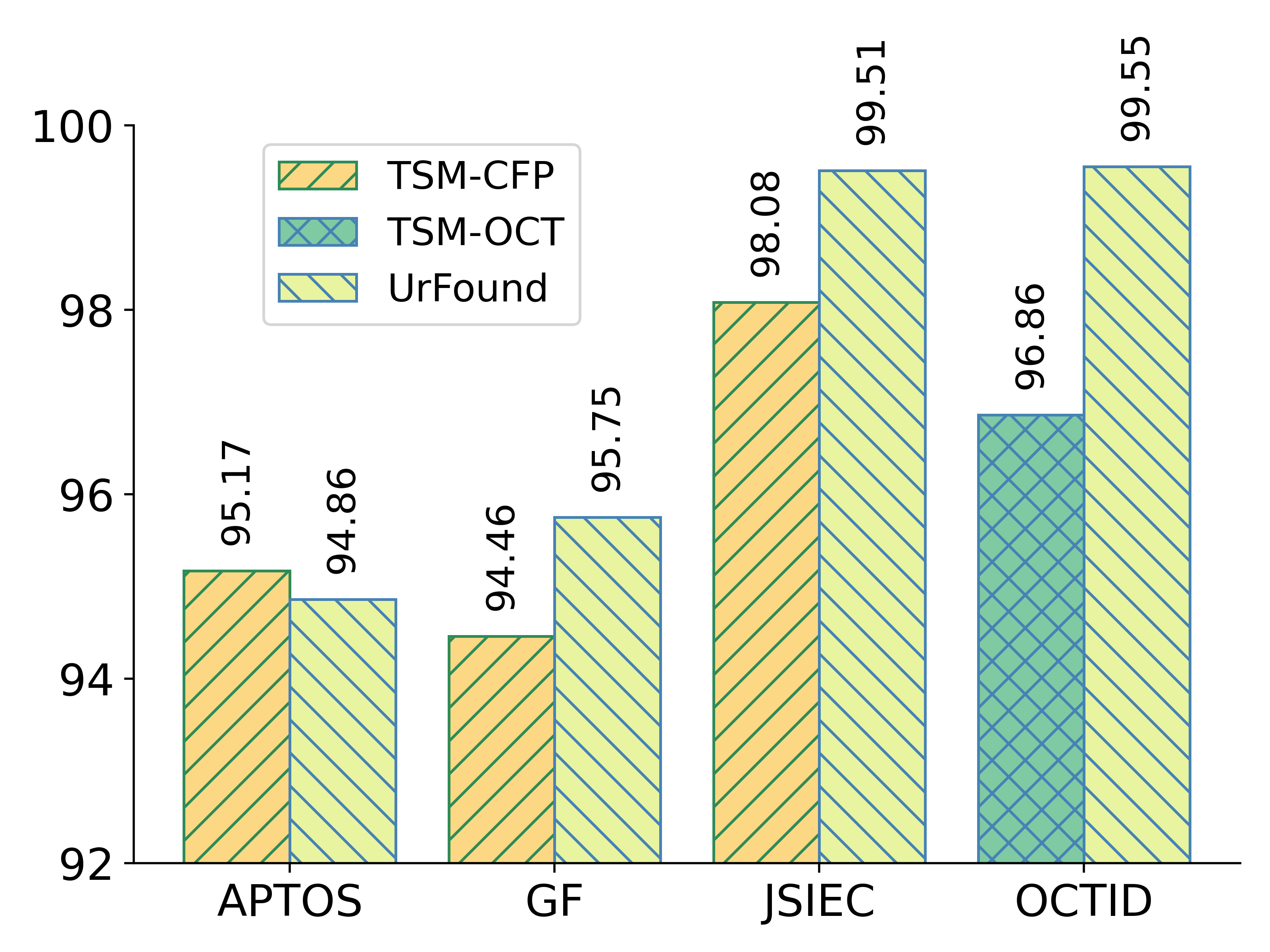}
            \put(-81,-3){\tiny{Dataset}}
            \put(-140,40){\rotatebox{90}{\tiny{AUROC}}}
        
        \caption{}
        \label{fig:sub_tsm}
    \end{subfigure}
    
    \caption{(a) Data efficiency of retinal FMs, Axes X and Y are the percentage of training data used and the corresponding AUROC, respectively. (b) Comparison of UrFound with tasks-specific models in AUROC with different datasets.}
\end{figure}

\textbf{Data efficiency.}
Fig. \ref{fig:sub_effi} show the classification results of the compared FMs at different percentages of training data on the APTOS, GF, JEIEC, and OCTID datasets. UrFound outperforms other retinal FMs in most settings and demonstrates a more significant advantage when fewer data are used for training (\textbf{Q4}). It is noteworthy that UrFound is pre-trained on $\sim$180k retinal images, a significantly smaller dataset compared to existing retinal FMs such as RETFound-CFP, which is trained with over 900k CFP images. These demonstrate the superior data efficiency of UrFound, making it well-suited for retinal imaging analysis with limited annotations.

\textbf{Comparison with task-specific models.}
To verify the advantages of transforming expert annotations into text for pre-training, we compare UrFound with task-specific models (TSMs) that are first trained with the class labels of specific classification tasks (as a supervised way of pre-training) and then adapted to test datasets for evaluation. Specifically, we test two TSMs: one for diabetic retinopathy grading (TSM-CFP) and the other for OCT disease classification (TSM-OCT). TSM-CFP is trained on all the CFP pre-training datasets for diabetic retinopathy grading, comprising 51,556 CFP images and labels of five classes. TSM-OCT is trained on all the OCT pre-training datasets, which include 83,484 OCT images and labels of four classes. In total, the data used for training TSMs account for 72\% of those used for pre-training UrFound.

Fig. \ref{fig:sub_tsm} presents the AUROC scores of TSMs and UrFound on the APTOS, GF, JSIEC, and OCTID datasets, where each dataset corresponds to different downstream tasks. UrFound and TSM-CFP obtain similar results on the ATPOS dataset. This is expected because the task of APTOS aligns with the training of TSM-CPT. UrFound consistently outperforms TSMs on other datasets. This suggests that TSMs lack the flexibility to learn generalizable representations for various tasks. In contrast, UrFound benefits from expert annotations via text supervision, offering a more effective approach to integrating valuable domain knowledge in representation learning (\textbf{Q5}).

\section{Conclusion}
We proposed UrFound, a \textbf{U}niversal \textbf{r}etinal \textbf{Found}ation model, which features a modality-agnostic image encoder and utilizes knowledge-guided mask modeling as a pre-training objective, allowing it to learn generalizable representations from both multimodal images and expert annotations. Through comprehensive experiments on 8 public retinal datasets, we demonstrated its strong generalization ability and data efficiency in adapting to various downstream tasks. Nevertheless, UrFound has two limitations: 
1. UrFound is designed to process CFP and OCT images while there exist other retinal imaging modalities such as FFA. 
2. UrFound is pre-trained on a relatively small dataset with disease labels as expert annotations. In practice, many unlabeled data are available for pre-training. 

\subsubsection{\ackname} This work was supported in part by the National Research Foundation of Singapore under its AI Singapore Programme (AISG) under Award AISG2-TC-2021-003, and in part by the Agency for Science, Technology and Research (A*STAR) through its AME Programmatic Funding Scheme under Project A20H4b0141, the RIE2020 Health and Biomedical Sciences (HBMS) Industry Alignment Fund Pre-Positioning (IAF-PP) (grant no. H20C6a0032), the 2022 Horizontal Technology Coordinating Office Seed Fund (Biomedical Engineering Programme – BEP RUN 3, grant no. C221318005) and partially supported by A*STAR Central Research Fund ``A Secure and Privacy-Preserving AI Platform for Digital Health''. 

\vspace{\baselineskip}
\noindent\textbf{Disclosure of Interests.} The authors have no competing interests to declare that are relevant to the content of this article.

\bibliographystyle{splncs04}
\bibliography{ref}


\newpage
\setcounter{section}{0}
\section{Supplementary Material}
\setcounter{figure}{0}
\setcounter{table}{0}

\begin{figure}[h]
\centering
\includegraphics[width=0.7\textwidth]{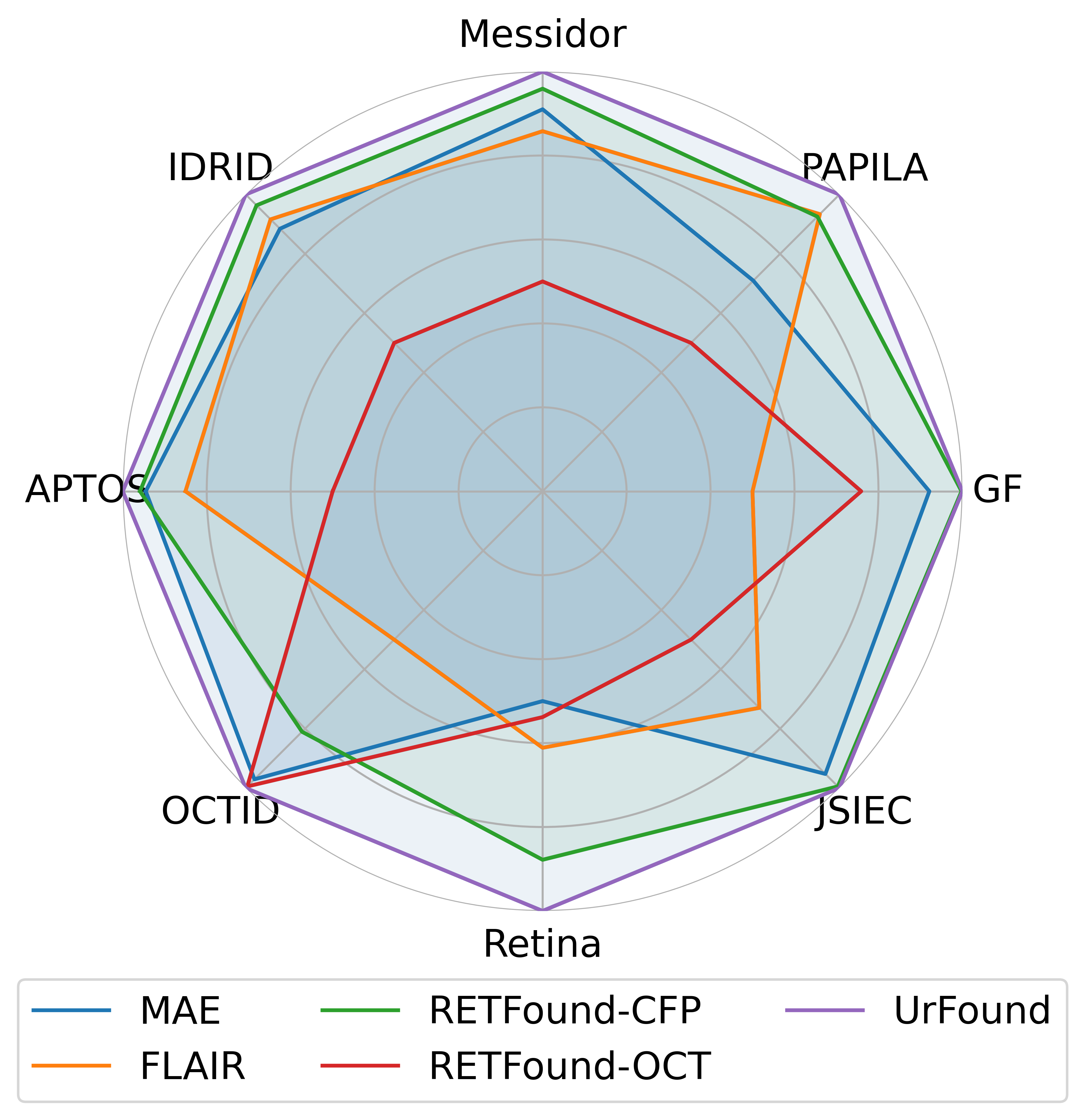} 
\caption{Performance of retinal FMs in adapting to different tasks.}
\label{fig:radar}
\end{figure}

\subsection{Dataset Preparation}

\noindent \textbf{Pre-training Dataset.}
Based on FLAIR~\cite{silva2023foundation}, we collected a large dataset (Tabel. \ref{app_tb1}) comprising 187,270 publicly accessible CFP and OCT images for the pre-training of our foundation model and the experiments conducted. Specifically, the OCT subset, sourced from the OCTCELL database \cite{kermany2018identifying}, includes 83,484 images, categorized under four distinct pathological labels: CNV (Choroidal Neovascularization), DME (Diabetic Macular Edema), DRUSEN, and NORMAL. Meanwhile, the CFP subset, compiled from 25 public datasets, consists of 103,786 images, extensively covering a wide array of retinal diseases. 

\begin{table}[t]
\centering
\caption{Collected publicly available dataset for foundation model pre-training.}
\label{app_tb1}
{\scriptsize
\begin{tabular}{clcp{8cm}}
\toprule
\textbf{No.} & \textbf{Name} & \textbf{Count} & \textbf{Labels} \\ 
\midrule
1 & OCTCELL~\cite{kermany2018identifying} & 83,484 & CNV, DME, DRUSEN, and NORMAL \\ 
2 & EYEPACS\cite{diabetic-retinopathy-detection} & 35,126 & noDR, mildDR, modDR, sevDR, prolDR \\ 
3 & RFMid~\cite{pachade2021retinal} & 3,170 & DR, ARMD, MH, DN, MYA, BRVO, TSLN, ERM, LS, MS CSR, ODC, CRVO, TV, AH, ODP, ODE, ST, AION, PT, RT RS, CRS, EX, RPEC, MHL, RP, CWS, CB, ODM, PRH, MNF, HR, CRAO, TD, CME, PTCR, CF, VH, MCA VS, BRAO, PLQ, HPED, CL \\ 
4 & EYENET~\cite{huang2021deepopht} & 15,709 & \textit{Text} \\ 
5 & LAG~\cite{li2019attention} & 4,854 & G, noG \\ 
6 & ODIR~\cite{odir2019} & 10,846 & N, DR, G, CAT, ARMD, HR, MYA \\ 
7 & PARAGUAY~\cite{benitez2021dataset} & 757 & noDR, mildDR, modDR, sevDR, prolDR \\ 
8 & STARE~\cite{hoover2000locating} & 397 & \textit{Text} \\ 
9 & ARIA~\cite{farnell2008enhancement} & 143 & N, ARMD, DR \\ 
10 & AGAR300~\cite{derwin2020novel} & 28 & DR, MA \\ 
11 & FUND-OCT~\cite{hassan2020rag} & 179 & G, N, CME, neovARMD, geoARMD, acCSR, chCSR \\ 
12 & DRIONS-DB~\cite{carmona2008identification} & 110 & noCAT, Dis \\ 
13 & Drishti-GS1~\cite{sivaswamy2014drishti} & 101 & N, G \\ 
14 & E-ophta~\cite{decenciere2013teleophta} & 265 & EX, MA \\ 
15 & G1020~\cite{bajwa2020g1020} & 1,020 & G, N \\ 
16 & HRF~\cite{budai2013robust} & 45 & N, G, DR, noisy \\ 
17 & ORIGA~\cite{zhang2010origa} & 650 & G, noG \\ 
18 & ROC~\cite{niemeijer2009retinopathy} & 100 & MA \\ 
19 & OIA-DDR~\cite{li2019diagnostic} & 13,673 & noDR, mildDR, modDR, sevDR, prolDR, HE, hEX, sEX, MA \\ 
20 & SYSU~\cite{lin2020sustech} & 1,219 & noDR, mildDR, modDR, sevDR, prolDR, HE, hEX, sEX \\ 
21 & JICHI~\cite{takahashi2017applying} & 9,939 & noDR, mildDR, modDR, sevDR, prolDR \\ 
22 & CHAKSU~\cite{kumar2023chakṣu} & 284 & G, noG \\ 
23 & DR1-2~\cite{pires2014advancing} & 1,469 & N, ReSD, hEX, DN, CWS, supHE, deepHE \\ 
24 & ScarDat~\cite{wei2019laser} & 997 & LS, noLS \\ 
25 & ACRIMA~\cite{diaz2019cnns} & 705 & G, noG \\ 
26 & DeepDRiD~\cite{liu2022deepdrid} & 2,000 & noDR, mildDR, modDR, sevDR, prolDR \\ 
\midrule
\multicolumn{2}{c}{\textbf{Total}} & 187,270 & \\ 
\bottomrule
\end{tabular}
}
\end{table}

\noindent \textbf{Categories.}
Here, based on FLAIR~\cite{silva2023foundation}, we provide the abbreviations for retinal diseases from our pre-training dataset along with their corresponding full names. Asteroid hyalosis (AH), Anterior ischemic optic neuropathy (AION), Age-related macular degeneration (ARMD), Branch retinal artery occlusion (BRAO), Branch retinal vein occlusion (BRVO), Cataract (CAT), Colobomas (CB), Choroidal folds (CF), Collateral (CL), Cystoid macular edema (CME), Choroidal neovascularization (CNV), Central retinal artery occlusion (CRAO), Chorioretinitis (CRS), Central retinal vein occlusion (CRVO), Central serous retinopathy (CSR), Cotton wool spots (CWS), Diabetic macular edema (DME), Drusen (DN), Diabetic retinopathy (DR), Drusen (DRUSEN), Disease (Dis), Epiretinal membrane (ERM), Exudate (EX), Glaucoma (G), Haemorrhages (HE), Haemorrhagic pigment epithelial detachment (HPED), Hypertensive retinopathy (HR), Laser scar (LS), Microaneurysms (MA), Macroaneurysm (MCA), Media haze (MH), Macular hole (MHL), Myelinated nerve fibers (MNF), Macular scar (MS), Pathologic myopia (MYA), Normal (N), NORMAL (Normal), Optic disc cupping (ODC), Optic disc edema (ODE), Optic disc pit maculopathy (ODM), Optic disc pallor (ODP), Plaque (PLQ), Preretinal haemorrhage (PRH), Parafoveal telangiectasia (PT), Post traumatic choroidal rupture (PTCR), Retinitis pigmentosa (RP), Retinal pigment epithelium changes (RPEC), Retinitis (RS), Retinal traction (RT), Red small dots (ReSD), Shunt (ST), Tilted disc (TD), Tessellation (TSLN), Tortuous vessels (TV), Vitreous haemorrhage (VH), Vasculitis (VS), Acute central serous retinopathy (acCSR), Chronic central serous retinopathy (chCSR), Deep haemorrhage (deepHE), Geographical age-related macular degeneration (geoARMD), Hard exudate (hEX), Mild diabetic retinopathy (mildDR), Moderate diabetic retinopathy (modDR), Neovascular age-related macular degeneration (neovARMD), No cataract (noCAT), No diabetic retinopathy (noDR), No glaucoma (noG), No large optic cup (noLS), Proliferative diabetic retinopathy (prolDR), Soft exudate (sEX), Severe diabetic retinopathy (sevDR), Superficial haemorrhage (supHE).

\noindent \textbf{Fine-tuning Dataset.}
To conduct a comprehensive evaluation of the foundation model, we collected 7 CFP datasets and 1 OCT dataset according to the experimental setup defined by RETFound~\cite{zhou2023foundation}, and divided them following the data division ratios provided by RETFound. As illustrated in Tabel. \ref{app_tb2}, APTOS, IDRID, and Messidor are utilized for the DR classification task based on CFP; PAPILA and GF are employed for the glaucoma classification task based on CFP; JSIEC and Retina are used for the multi-disease classification task based on CFP; and OCTID is designated for the multi-disease classification task based on OCT.

\begin{table}[t]
\centering
\caption{Collected publicly available dataset for foundation model fine-training.}
\label{app_tb2}
{\scriptsize
\begin{tabular}
{C{2.6cm}|C{1.2cm}|C{1.2cm}C{1.2cm}C{1.2cm}|C{1.2cm}}
\toprule
\textbf{Task} & \textbf{Dataset} & \textbf{Train} & \textbf{Val} & \textbf{Test} & \textbf{Total}\\ 
\midrule
\multirow{3}{*}{\textbf{DR}} & APTOS & 2,048 & 514 & 1,100 & 3,662 \\
 & IDRID & 329 & 84 & 103 & 516 \\
 & Messidor & 972 & 264 & 526 & 1,744 \\
\midrule
\multirow{2}{*}{\textbf{Glaucoma}} & PAPILA & 313 & 77 & 98 & 488 \\
 & GF & 861 & 218 & 465 & 1,544 \\
\midrule
\multirow{3}{*}{\textbf{Multiple Category}} & JSIEC & 534 & 150 & 313 & 997 \\
 & Retina & 336 & 84 & 181 & 601 \\
 & OCTID & 317 & 80 & 175 & 572 \\
\bottomrule
\end{tabular}
}
\end{table}

\noindent \textbf{Task Specific Dataset.}
Based on the labels in the pre-training dataset, we constructed a task-specific dataset for Diabetic Retinopathy classification, which includes images from EYEPACS, PARAGUAY, OIA-DDR, and DeepDRiD, totaling 51,556 images. Similarly, a task-specific dataset for OCT disease classification was developed based on the OCTCELL dataset.

\subsection{Expert Knowledge Descriptions.}
For the domain knowledge descriptors related to retinal diseases based on CFP, we referred to FLAIR~\cite{silva2023foundation} for guidance. Meanwhile, for the domain knowledge descriptors concerning retinal diseases based on OCT, we utilized ChatGPT-4 to summarize four distinct descriptions for the corresponding disease label names, which were then employed as the domain knowledge descriptors (Tabel. \ref{app_tb3}).

\begin{table}[t]
\centering
\caption{Expert Knowledge descriptions for OCT-based retinal diseases.}
\label{app_tb3}
{\scriptsize
\begin{tabular}
{C{2cm}|p{10cm}}
\toprule
\textbf{Category} & \textbf{Domain Knowledge descriptor} \\ 
\midrule
\multirow{4}{*}{\textbf{CNV}} & 1. ``The OCT image reveals a network of new blood vessels beneath the retinal pigment epithelium, indicative of Choroidal Neovascularization. These vessels are irregular and often associated with age-related macular degeneration." \\
 & 2. ``There is noticeable distortion and elevation of the overlying retinal layers, which is characteristic of the leakage and bleeding from these abnormal vessels." \\
 & 3. ``Pockets of fluid accumulation under the retina, known as subretinal fluid, are evident, causing a dome-shaped elevation of the retina." \\
 & 4. ``Areas of hemorrhage and exudation are visible between the retinal layers and beneath the retinal pigment epithelium, indicating active vascular leakage." \\
\midrule

\multirow{4}{*}{\textbf{DME}} & 1. ``In Diabetic Macular Edema, the OCT scan shows a significant thickening of the macula, particularly in the inner retinal layers, due to fluid accumulation. This condition is a common complication of diabetic retinopathy." \\
 & 2. ``Multiple cystic spaces within the retinal layers are observed, filled with fluid, giving a sponge-like appearance to the retina." \\
 & 3. ``Hyperreflective foci are seen below the retinal pigment epithelium, representing hard exudates, which are residues of lipid deposits from leaking blood vessels." \\
 & 4. ``In advanced cases, disruption and irregularity of the retinal pigment epithelium layer are noted, likely due to chronic edema and vascular leakage." \\
\midrule

\multirow{4}{*}{\textbf{DRUSEN}} & 1. ``Drusen appear as small, round elevations beneath the retinal pigment epithelium layer in OCT images. These are accumulations of extracellular material, commonly associated with age-related macular degeneration." \\
 & 2. ``The drusen vary in size and confluence, with larger and more numerous drusen indicating a higher risk of progression to advanced macular degeneration." \\
 & 3. ``In cases of extensive drusen, there is noticeable distortion and thickening of the overlying retinal pigment epithelium layer." \\
 & 4. ``Some drusen exhibit a central hyperreflective core with a surrounding hyporeflective halo, suggesting varying stages of drusen evolution." \\
\midrule

\multirow{4}{*}{\textbf{NORMAL}} & 1. ``The normal retina in OCT imaging presents a well-defined, multi-layered structure. Each layer exhibits its characteristic reflectivity, with clear demarcation between layers." \\
 & 2. ``The retinal pigment epithelium layer appears as a uniform, thin band adjacent to the highly reflective Bruch's membrane." \\
 & 3. ``The photoreceptor layer, including the cones and rods, is orderly and shows no signs of fluid accumulation or structural distortion." \\
 & 4. ``The nerve fiber layer, ganglion cell layer, and inner and outer nuclear layers all display normal thickness and reflectivity, with no signs of pathology or abnormality." \\
 
\bottomrule
\end{tabular}
}
\end{table}

\end{document}